# Machine Learning Based Optimal Design of Fibrillar Adhesives

Mohammad Shojaeifard[1], Matteo Ferraresso[1], Alessandro Lucantonio[2,+] Mattia Bacca[1,*]

[1] *Mechanical Engineering Department, University of British Columbia, Vancouver BC V6T1Z4, Canada*
[2] *Department of Mechanical and Production Engineering, Aarhus University, Aarhus, Denmark*

**Abstract**

Fibrillar adhesion, observed in animals like beetles, spiders, and geckos, relies on nanoscopic or microscopic fibrils to enhance surface adhesion via 'contact splitting.' This concept has inspired engineering applications across robotics, transportation, and medicine. Recent studies suggest that functional grading of fibril properties can improve adhesion, but this is a complex design challenge that has only been explored in simplified geometries. While machine learning (ML) has gained traction in adhesive design, no previous attempts have targeted fibril-array scale optimization. In this study, we propose an ML-based tool that optimizes the distribution of fibril compliance to maximize adhesive strength. Our tool, featuring two neural networks (NNs), recovers previous design results for simple geometries and introduces novel solutions for complex configurations. The Predictor NN estimates adhesive strength based on random compliance distributions, while the Designer NN optimizes compliance for maximum strength using gradient-based optimization. Our method significantly reduces test error and accelerates the optimization process, offering a high-performance solution for designing fibrillar adhesives and micro-architected materials aimed at fracture resistance by achieving equal load sharing (ELS).

Corresponding authors: [*]mbacca@mech.ubc.ca; [+]a.lucantonio@mpe.au.dk

## Introduction

Fibrillar structures are widespread in nature, and are particularly observed at the terminal part of animal limbs that have the ability to attach to rough surfaces to defeat gravity. This is the case for beetles, spiders, and the gecko [1-3]. These dry adhesive structures enable robust, reversible, and repeatable adhesion without the need for chemical depletion, thus making them ideal candidates for bioinspired adhesive devices. Engineering applications span all fields, such as robotics [4], wearable electronics [5-6], medicine [7-8], 3D printing [9], and mechanical manipulators [10-11]. Studies have shown that smaller, finely divided terminal fibrillar structures, composed of adhesive units that we call 'fibrils', enhance the adhesive strength by increasing effective contact surface area of the interface. Adhesive strength has been shown to inversely scale with fibril size and align with the Johnson, Kendall, and Roberts' (JKR) theory [12]. Notably, as fibrils get smaller and adhesive strength increases, one can observe a transition to a strength saturation regime [13]. In this regime, stress concentrations are no longer controlling detachment, and we achieve 'flaw insensitivity', thus, giving a cohesive detachment mechanism. Here, all contact points detach simultaneously once the interfacial stress trespasses a theoretical limit in adhesive strength [13-14]. This condition, defined as 'equal load sharing' (ELS), provides then the maximum theoretical adhesion strength of the interface, which is independent of geometry and only depends on the chemical composition of the materials in contact. While producing smaller and smaller fibrils is a viable strategy in nature to achieve ELS; this is challenging, if not impossible, in engineering prototypes. Thus, a need to search for alternative strategies to maximize adhesion. Another natural strategy to homogenize interfacial stresses, and thus reduce stress concentrations, is the use of functional grading of physical properties. Examples in nature have shown that functional grading of the elastic modulus across the length of the single fibril can reduce stress concentration favoring ELS [15-16]. Here, soft tips and rigid stalks not only promote interfacial stress uniformity but also prevent fibril condensation (self-adhesion), a known detrimental effect in engineering prototypes [17]. Other engineering examples and computations have shown the benefits of functional grading of the elastic modulus across the surface of the sample to reduce stress concentrations [18-20]. Other design strategies aimed at improving the adhesive strength of the single unit, is the use of defect-tolerant geometries like mushroom-shaped fibrils [21]. To further advance the exploration of optimal fibril geometry and elastic modulus distribution, recent studies have explored the use of techniques like topology optimization [20], deep learning [22], and machine learning (ML) [23-24]. The abovementioned attempts to improve the adhesive strength are limited to that of the single adhesive unit. As shown by [25-29], the adhesive strength of the whole interface is commonly much smaller than that of the single fibril. This is due to inter-fibril stress concentrations generated by a compliant backing layer (substrate from which the fibrils protrude) and by an imperfect alignment between the adhesive and the target surface. Thus, to achieve ELS across the interface, one must adopt design strategies to reduce such stress concentrations so that all fibrils carry the same load at detachment. As proposed by [27,29], functional grading of fibril compliance across the interface, where peripheral fibrils -commonly carrying larger loads- are made more compliant than central fibrils, has resulted in producing ELS. However, the specific distribution of fibril compliance was first defined numerically [27] and then analytically, via asymptotic solution [29], only for selected array shapes like square and circular. For a generic array shape, the numerical solution provided by [27] might provide valuable results but at a significant computational cost. In this paper we propose to use ML to define the optimal functional grading of fibrillar interfaces to achieve maximum adhesive strength (and thus ELS). This provides a powerful design tool for adhesive interfaces, which can extend well beyond fibrillar dry adhesion. Our ML-based optimization framework is based on an inverse design approach, specifically developed to determine the optimal compliance distribution within a fibrillar array, leading to the highest adhesion performance. This framework leverages two deep neural networks, the *predictor*, which provides design principles, and the *designer*, which is trained on existing datasets from simulation results of interfacial detachment. The architecture and learning parameters derived from this training are then utilized to identify the optimal inverse design. Our ML-based approach employs backpropagation to compute the analytical gradients of the objective function concerning the design variables. It effectively avoids local minima traps by enabling rapid gradient calculations and conducting multiple optimizations with different initial values. This framework is capable of optimizing adhesion strength across any fibrillar array configuration, achieving results that are otherwise difficult to obtain using analytical solutions [29]. Our results are finally compared to the previously obtained solution [27,29], showing good agreement. We also analyze several optima provided by our ML-based tool and show how several solutions, which might be more practically attainable, give adhesive strength that is sufficiently close to the theoretical maximum given at ELS.

**Mechanical model for fibrillar adhesion**

Several studies investigated fibrillar adhesion and the condition of equal load sharing (ELS) among fibrils across the interface [25-29]. A recent investigation [27] provided a theoretical model to describe fibrillar adhesion considering backing layer (BL) interactions and their effect on ELS. This model considers an array of $N$ adhesive units, the fibrils, which are positioned atop an elastic half-space, the BL. Both fibrils and BL are made of a homogenous and isotropic linear elastic material. Figure 1a provides the schematics of our model system, where the fibrils are packed with orthogonal distribution along $x$ and $y$-axes, with a fixed center-to-center distance $l$. Each fibril's tip is in contact with a rigid surface (RS) to which they transmit contact adhesive forces. The generic fibril $i$ is therefore loaded with the tensile force $f_i$. The total force exchanged between RS and BL is the sum of all tensile loads carried by the fibrils via

$$F = \sum_{i=1}^{N} f_i \tag{1a}$$

subjected to the condition

$$f_i < f_c \tag{1b}$$

where violation of Eq. (1b) simply discounts fibril $i$ from the force balance in (1a). Eq. (1b) is violated when $f_i = f_c$ and this occurs when fibril $i$ detaches from the RS ($f_c$ being the critical detachment load of each fibril). From Eq. (1) we can immediately see that $F \leq Nf_c$, where $Nf_c$ represents the maximum theoretical adhesive strength describing ELS, and occurring only with the simultaneous detachment of all fibrils.

We perform our analysis in displacement control, by prescribing the distance $\bar{u}$ between the RS and the BL and their misalignment angles $\theta_{xz}$ and $\theta_{yz}$, along both $x$ and $y$-axes, giving [27]

$$d_i = \bar{d} + \tan\theta_{xz}\, x_i + \tan\theta_{yz}\, y_i \tag{2}$$

as the deflection of the tip of each generic $i$–th fibril. In Eq. (2), $x_i$ and $y_i$ are the cartesian coordinates of fibril $i$, function of the center-to-center distance $l$, and $\bar{d}$ provides the RS-BL separation distance at $x = y = 0$. In the case of perfect RS-BL alignment we have $\theta_{xz} = \theta_{yz} = 0$, so Eq. (2) simplifies to $d_i = \bar{d}$, where all fibril tip displaces by the same prescribed amount.

To connect fibril $i$'s deflection $d_i$ with its carried tensile load $f_i$ we invoke linear elastic contact mechanics [30], following the relation provided by [27]

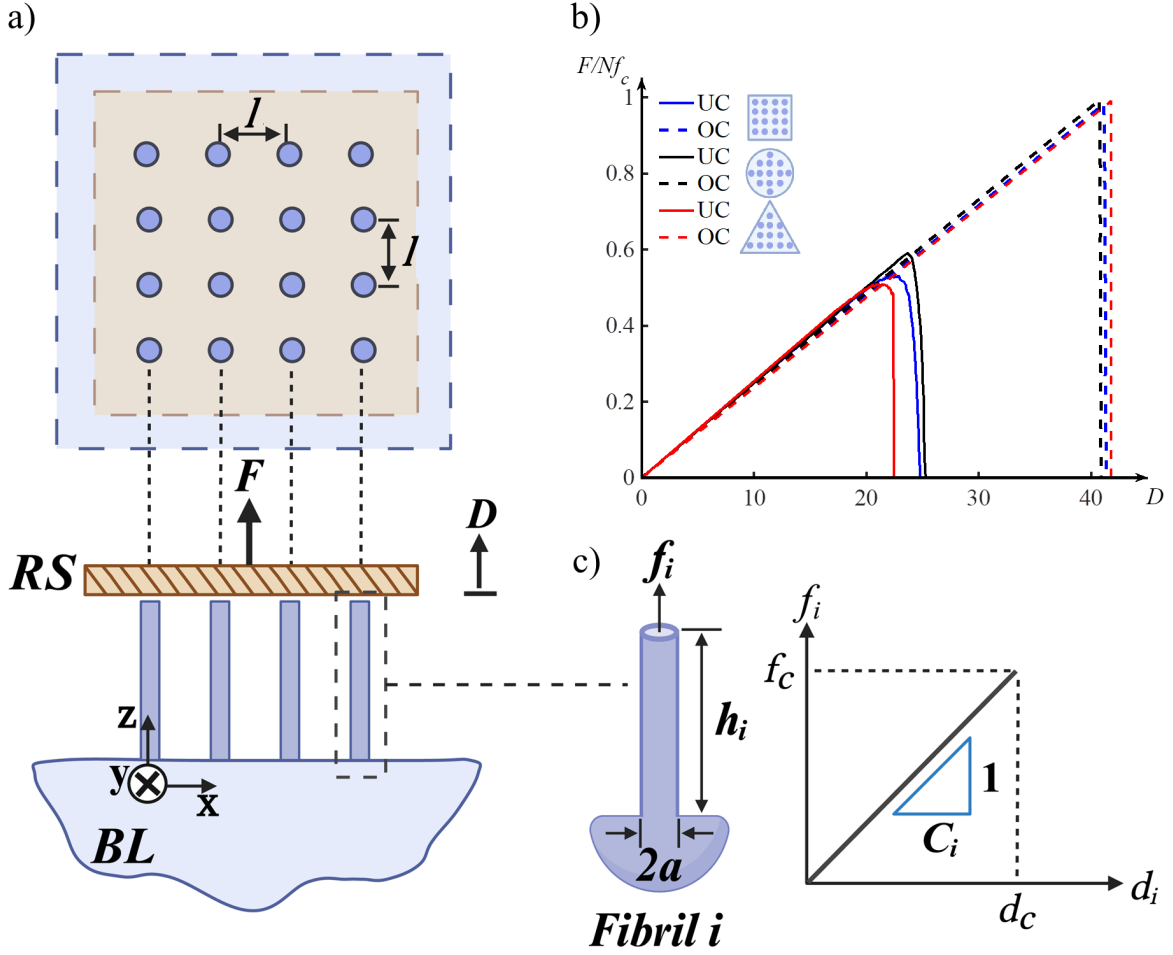

**Figure 1**: Schematics of our model system. (*a*) The adhesive is composed of a square array of cylindrical fibrils mounted on atop an elastic half space depicting the backing layer (BL). The fibrils are in adhesive contact with the rigid surface (RS). (*b*) Macroscopic force-deflection behavior at the interface, where the dimensionless force is $F/Nf_c$ and the dimensionless deflection is $D$ (Eq. (4d) and (6)). The adhesive interface experiences a force peak $F_c$ during detachment. This is due to the competition between elastic loading and the progressive detachment of fibrils. Fibrillar arrays composed of uniform conditions (UC) among fibrils (solid lines) exhibit suboptimal strength, where $F_c < Nf_c$; while adhesives composed of optimally-graded conditions (OC) [27,29] provide optimal strength, giving $F_c = Nf_c$. The latter is due to the optimal distribution of fibril compliance $C_i$, describing the load-deflection behavior of the single fibril $i$. (c) Schematics of force-deflection ($f_i - d_i$) of the single fibril, highlighting the fibril compliance $C_i = d_i/f_i$ subject to optimal distribution (functional grading), as well as the 'brittle' behavior caused by the sudden detachment when $f_i = f_c$. In the reported results we adopted the cross-sectional radius $a_i = a$, length $h_i = 5a$, and center-to-center spacing $l_i = 3.57a$ for each fibril.

$$d_i = d_i^{fib} + \sum_{j=1}^{N} d_{ij}^{BL} \tag{3a}$$

Here,

$$d_i^{fib} = \frac{h_i}{\pi a_i^2 E_i} f_i \tag{3b}$$

is the tip displacement of fibril $i$ due to its linear elastic tensile strain in response to its load $f_i$,

$$d_{ij}^{BL} = \frac{1}{\pi\, r_{ij}\, E^*} f_j, \quad \text{for } i \neq j \tag{3c}$$

is that due to the deflection of the BL caused by the loads carried by the neighbor fibril $j$, and finally,

$$d_{ij}^{BL} = \frac{16}{3\,\pi^2\, a_i\, E^*} f_j, \quad \text{for } i = j \tag{3d}$$

is that due to the BL deflection at the attachment with fibril $i$ caused by the load carried by fibril $i$ itself. In Eq. (3), $h_i$, $a_i$, and $E_i$ are the length, cross-sectional radius, and elastic modulus or fibril $i$, $r_{ij}$ is the center-to-center distance between fibrils $i$ and $j$, and $E^* = E/(1-\nu^2)$ is the plane strain elastic modulus of the BL, where $E$ and $\nu$ are the elastic modulus and Poisson's ratio of the BL. In the case of an incompressible BL, we can simply write $E^* = 4E/3$.

Eq. (3a) can then be rewritten in dimensionless vectorial form [27] as

$$\frac{d_i}{d_0} = \sum_{j=1}^{N} C_{ij} \frac{f_j}{f_c} \tag{4a}$$

subjected to the condition at Eq. (1b) ($f_j/f_c < 1$), where

$$C_{ij} = \frac{a}{r_{ij}}, \quad \text{for } i \neq j \tag{4b}$$

is the compliance term due to BL interactions,

$$C_{ij} = \frac{16}{3\pi}\left(\frac{a}{a_i}\right) + C_i, \quad \text{for } i = j \tag{4c}$$

is the compliance associated with the single fibril's interaction with the BL (first term on the right-hand side) and the fibril's extension compliance (second term), with

$$C_i = \left(\frac{E^*}{E_i}\right)\left(\frac{a}{a_i}\right)^2\left(\frac{h_i}{a}\right) \tag{4d}$$

Also,

$$d_0 = \frac{f_c}{E^*\pi a} \tag{4e}$$

is the nominal displacement, and $a = \sum_{i=1}^{N} a_i/N$ is the average fibril cross-sectional radius, taken as a nominal length. Here we define the *dimensionless compliance tensor* as $\mathbf{C}$, which components are given in Eq. (4b) and (4c). The second term on the right-hand side of Eq. (4c) represents the design variable in each fibril, where one can functionally grade the dimensionless length $h_i/a$ or modulus $E_i/E^*$ in each fibril. Because we don't consider variations on the first term on the right-hand side of Eq. (4c), we exclude the possibility of tailoring $a_i$, however, this would be a simple extension to our model. We should also consider that a gradient in fibril height $h_i$ would produce compatibility challenges for a flat RS. Thus, the simplest design example following from our study would be to tailor the fibril modulus $E_i$.

By inverting Eq. (4a), we obtain the dimensionless load carried by each fibril as a function of their tip deflection

$$\frac{f_i}{f_c} = K_{ij} \frac{d_j}{d_0} \tag{5}$$

where the *dimensionless stiffness matrix*, $\mathbf{K}$, is the inverse of the compliance matrix, *i.e.*, $\mathbf{K} = \mathbf{C}^{-1}$. We prescribe RS-BL separation via the dimensionless deflection

$$D = \frac{\bar{d}}{d_0} \quad (6)$$

from an initial condition of $D = 0$, and further increments by a fixed $\Delta D$, so that we have $D^{k+1} = D^k + \Delta D$ at each $k$-th iterative step. During the analysis, at each step, we plot the dimensionless force $F/Nf_c$ required to achieve the prescribed separation $D$. Whenever the condition at Eq. (1b) is violated, *i.e.*, when $f_i/f_c = 1$, for a fibril $i$, we consider this fibril is detached and remove its contribution from Eq. (1) and (3a) or (4a). By removing the detached fibril's contribution, we observe a progressive reduction of the force-deflection ($F - D$) slope and the simulation continues until $F = 0$, as shown in Figure 1b. The competition between elastic loading of both fibrils and BL, and the 'interfacial damage' produced by the progressive detachment of overloaded fibrils generates a maximum (peak) force $F_c/Nf_c$, which describes the adhesive strength of the interface. As we can see from Figure 1b, the peak force $F_c/Nf_c$ is always smaller than unity, describing suboptimal strength for fibrillar interfaces that are sub-optimally designed, like in the case of an array with identical (non-functionally-graded) fibrils (solid lines). For the exceptional case of optimal functional grading (dashed lines), obtained from the solutions of [27,29], we achieve the optimal theoretical strength for which $F_c/Nf_c$ is unity ($F_c = Nf_c$), and recover ELS conditions. The force-deflection plots in Figure 1b are generating adopting a cross-sectional radius $a_i = a$, length $h_i = 5a$, and center-to-center spacing $l_i = 3.57a$ for each fibril. Finally, Figure 1c provides a schematic of the fibril compliance $C_i$, subject of our optimal grading.

**Machine Learning Framework**

Bacca et al. [27] suggest that a collection of fibrils can attain Equal Load Sharing (ELS) through an optimal distribution of compliance. This is achieved by having more compliant peripheral fibrils compared to the central ones, so to that all fibrils carry the same load at the onset of detachment, and thus detach simultaneously. ELS gives the interface its maximum theoretical adhesive strength, *i.e.*, the sum of the critical detachment load of all the fibrils. In this section we aim at obtaining the optimal compliance distribution among fibrils across the adhesive interface, giving ELS, using Inverse Design Networks (IDNs). We employed a variety of machine learning (ML) models and assessed their performance to identify the most efficient one that also provides the highest accuracy in the solution. Specifically, we trained Linear/Non-linear Regression models and Multilayer Perceptron (MLP) neural networks. By randomly distributing fibril compliance and then calculating the detachment strength with the model described in the previous section, we generated a dataset composing 2500 entries (datapoints). Each sample includes the distribution of fibril compliance as features, and the adhesive strength as a label. The 2500 adopted samples were specifically chosen among those having adhesive strength that is below 70% of the maximum. We made this selection to test our method in its ability to extrapolate an optimum from 'suboptimal' training data. To make fair comparisons among solutions, we fixed the average fibril compliance in all the 2500 simulations. Following common practices in the ML field, we randomly split the dataset into a training set (80%) and a test set (20%), using the latter to assess the predictive capability (generalization) of the models. To carry out model selection, the optimal values for hyperparameters such as the number of hidden layers and neurons that minimized the validation error were found using a grid search within a 5-fold cross-validation [32]. This method consists in splitting the training data into 5 folds, using 4 of them for training and 1 for validation; the procedure is repeated for the 5 possible choices of the validation set and the validation error is computed by averaging the results.

As reported in Table 1, the performance of 6-layer neural network, utilizing Mean Squared Error (MSE) between the estimated and the predicted forces as our performance metric, significantly surpassed other models in predicting the adhesive force of our fibrillar array.

**Table 1** Prediction accuracy and parameter values of the proposed ML models for training and test data.

| Model | Parameters | Training | | Test | |
|---|---|---|---|---|---|
| | | MSE | $R^2$ | MSE | $R^2$ |
| Linear Regression | With bias | 0.0152 | 0.601 | 0.0156 | 0.596 |
| Non-linear regression | Polynomial order =3 | 0.0131 | 0.65 | 0.0134 | 0.64 |
| Non-linear Gaussian RBF | | 0.0098 | 0.743 | 0.0105 | 0.730 |
| 1-layer MLP | # Neurons per layer =64 | 8.9e-5 | 0.998 | 7.4e-5 | 0.997 |
| 6-layer MLP | # Neurons per layer =64 | 1.5e-5 | 0.9996 | 1.8e-5 | 0.9995 |

Upon determining that the 6-layer MLP outperforms other models in predicting the adhesive force of a given adhesive array (and its compliance distribution), we integrated the network into our framework to determine the optimal compliance distribution across the fibril array. Consequently, as depicted in figure 2, our IDNs framework is composed of two NNs: one functioning as a *predictor* and the other one as a *designer*, both sharing an identical neural network architecture. The predictor within the IDNs framework operates as a forward predictive model, trained to closely approximate a predefined equation. The learning variables within the predictor NN consist of the weights and biases that form connections between neurons in successive layers. Once the predictor is adequately trained, these learned weights and biases, along with the preserved architecture of an equivalent number of layers and neurons, are transferred to the designer NN. Contrary to the predictor NN, the designer NN functions as an inverse design model where the weights and biases become fixed constants. In this model, the design variables themselves are established as the learning parameters. Therefore, the training procedure for the designer NN transforms into a design optimization process, with the primary goal of maximizing the desired characteristic, which is the adhesion force of the fibrillar array. Optimized designs are subsequently produced as outputs, derived from gradients computed through the backpropagation process. Within the feedback loop, the algorithm verifies the optimized designs, which may then be incorporated into the existing training data set, setting the stage for subsequent iterations of training and design refinement.

The code used for implementing the NN and running simulations is available on GitHub at https://github.com/MattiaBacca/Fibrillar-Adhesives-ML-based-Design. The repository contains all the information necessary to reproduce the results presented in this study.

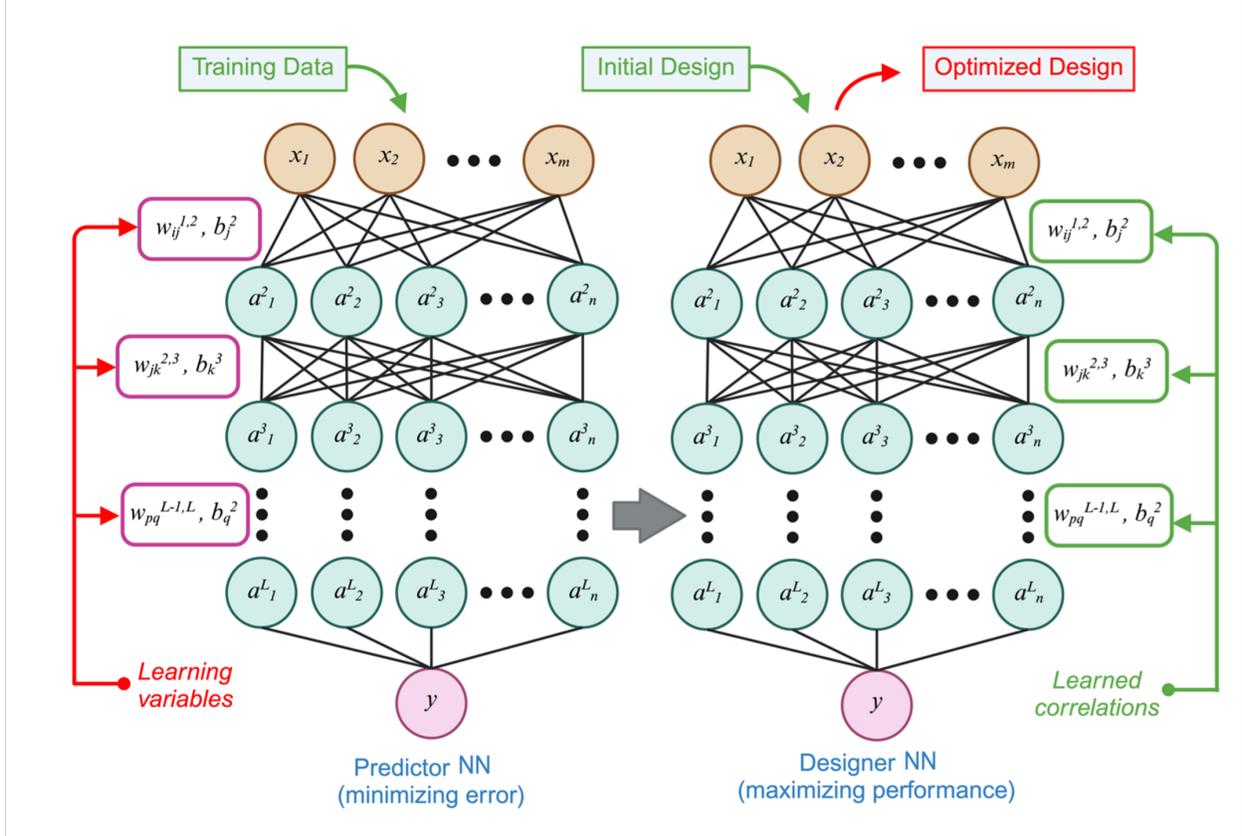

**Figure 2** The framework of Inverse Design Networks consists of the *predictor* and the *designer* NNs to, respectively, predict and optimize the adhesion strength of a fibrillar adhesive.

**Results and Discussion**.

In this section, we present the outcomes derived from Inverse Design Networks (IDN) framework, as illustrated in Figure 2. Our objective is to obtain the optimal fibril compliance distribution (design) that maximizes adhesive strength, *i.e.*, the maximum detachment force. By employing our inverse design approach, we systematically assessed various array configurations given by their compliance distribution across the fibril array. Informed by the NN robust performance reported in Table 1, we employed a subset of 2500 samples to train the Predictor NN. These samples include the compliance distribution and the associated adhesive strength, calculated from a numerical simulation (see the section 'Mechanical model for fibrillar adhesion'). The adhesive strength of a fibrillar array with uniform fibril compliance is a fraction of the maximum theoretical, and this fraction depends on the size of the array (number of fibrils) and the competition between backing layer (BL) compliance and fibril compliance [27-29]. An adhesive with rigid backing layer, or infinitely compliant fibrils, can achieve maximum strength (*i.e.*, the sum of the adhesive strength of all fibrils) [27]. Conversely, stiffer fibrils and/or compliant BL produce arrays with very low adhesive strength. This is due to the fact that load concentrations among fibrils are controlled by BL interaction, promoted by BL compliance. Considering 2500 samples with adhesive strength below 70% of the maximum allows us to evaluate our method's capability to extrapolate an optimum from suboptimal data.

Figure 3 illustrates the performance of our NN model by comparing the adhesion strength to its predicted value for both the training (left) and test (right) datasets. In both the reported plots, the real adhesive strength is in the *x*-axis, while the predicted one is in the *y*-axis. The identity line, $y = x$, is also plotted as a reference to the ideal 100% accuracy in the adhesive strength prediction. We also report two parallel blue dashed lines to indicate a tolerance of 3% error in the prediction. Here we can observe that our NN tends to overestimate adhesive strength, but never beyond the 3% tolerance, while nearly no underestimation is reported. The R-squared for the training set of our predictive model is $R^2 = 0.9996$, while that of the test set is $R^2 = 0.9995$, indicating very high accuracy.

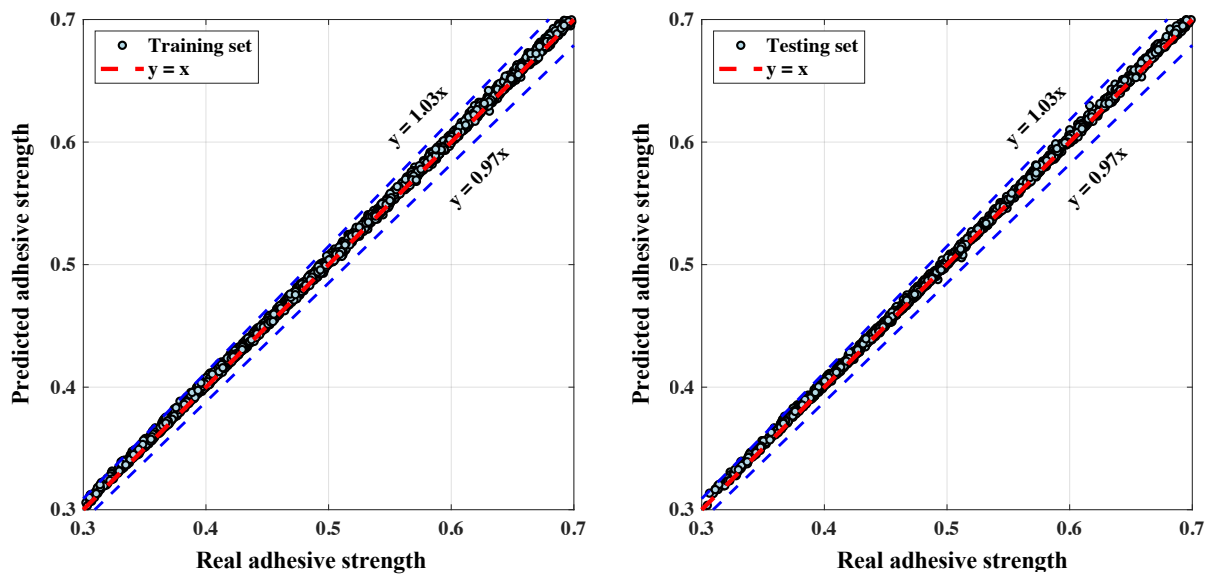

**Figure 3** Performance analysis of the proposed NN model for the training and test data set.

Once the training of the Predictor NN is completed, the architecture of the model, along with the calculated values of the weights and biases, are transferred to the Designer NN. Then, the designer NN is utilized to search for the global maximum by using backpropagation starting with 100 random initial points. While backpropagation, as utilized in the Designer NN, is a gradient-based optimization method that could potentially converge to local maxima, it calculates analytical gradients, offering a significant advantage in computational speed and efficiency over alternative numerical techniques such as the finite-difference methods. Upon solving the inverse design with 100 varied initial conditions, numerous solution paths were observed to converge on the global maximum. This suggests that employing our strategy of random initializations for the design variables is an effective approach to mitigate the issue of local maxima. As follows, we analyze some specific geometry of fibrillar arrays and explore the optimal distribution of compliance to maximize the adhesive strength.

First, we analyze a circular array of fibrils. This is characterized by a radius $R = 75a$, with $a$ the radius of the fibril (Figure 1), a fibril center-to-center spacing $d = 3a$ and a fibril Young's modulus $E_f = E$, with $E$ the modulus of the BL, and finally a Poisson's ratio $v = 0.5$ in the BL.

We obtain the optimal dimensionless compliance distribution as a function of the normalized fibril position $r/R$ and report it in Figure 4. This figure illustrates the top five compliance distributions that yield the highest adhesion strengths as calculated through the machine learning (ML) optimization framework using various initializations. The dimensionless adhesion strength (normalized by the theoretical maximum) obtained with these solutions is 0.984, 0.977, 0.973, 0.968, and 0.942, from the 1st to the 5th best solutions. The proximity of these dimensionless strength to unity proves the effectiveness of our method. As a benchmark, the fibrillar array having homogeneous distribution of fibril compliance (equal to the average compliance) gives a dimensionless strength of 0.58. Figure 4-*left* shows that the provided solutions are close to the analytical optimal solution from [29] for $r/R > 0.6$, *i.e.*, sufficiently far from the center of the array, while they digress from the theoretical benchmark for $r/R < 0.6$ with a discrepancy that increases from the 1st to the 5th solution. In Figure 4-*right* we report the adhesive strength of all five solutions, ordered form 1st to 5th. The four best solutions have minimal discrepancy in strength among themselves and from the theoretical benchmark, while the 5th solution shows a more remarked drop in strength. This solution is also the one that shows the highest discrepancy in compliance distribution in the center of the array. Note that the fibril compliance $C$ ($C_i$ for fibril *i*) reported in the y-axis of Figure 4-*left* is not normalized by the average compliance but by instead is normalized as $(C - C_{min})/(C_{max} - C_{min})$, so that all four curves in this figure match the maximum at the periphery. Thus, the area underneath each curve in this figure is not equal. In the case of ML Opt 5, the 5th best solution, the higher area underneath the curve suggests that the maximum compliance at $r = R$ is smaller than that of the other curves, and hence its reduced performance. Nevertheless, the significant deviation of ML Opt 5, compared to the theoretical benchmark [29], at the center of the array ($r < 0.6\ R$) suggests that the peripheral fibrils play a bigger role in defining the optimal performance. This is also likely due to the larger portion of surface contained in the outer ring of the circular adhesive.

Next, we explore a square fibrillar array. The square array has the same outer dimensions as the circle in the previous example (diameter of the circle is equal to the side of the square). In this case we also have the same fibril spacing and elastic properties of both fibrils and BL as in the previous example. In Figure 5 we report the obtained five best solutions (in fibril compliance distribution) giving the highest adhesion strength. As observed in the previous example, and Figure 4, the optimum design involves softer fibrils at the perimeter and stiffer ones at the center. As a benchmark of comparison, the array having uninform compliance distribution gives a dimensionless adhesion strength, normalized by the theoretical maximum, of 0.53. Also in this case, as in Figure 4, we observe a close proximity in compliance distribution, for all the five best solution, for $r/R > 0.6$, while the biggest discrepancies emerge at the enter of the array for $r/R < 0.6$. Once again, the minor difference in adhesive strength among solution suggests the key role of peripheral fibrils.

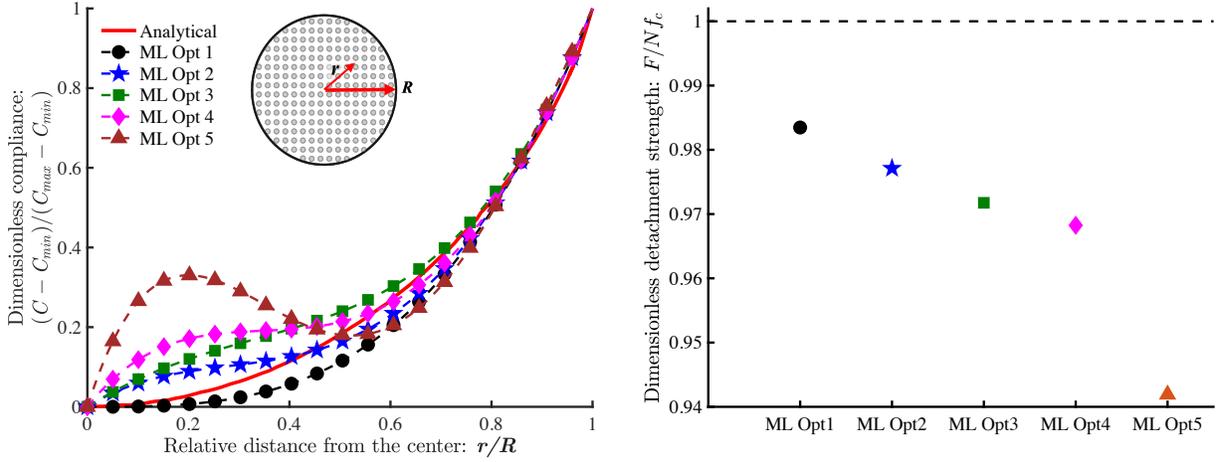

**Figure 4** Optimal fibril compliance distribution (*left*) and corresponding adhesion (detachment) strength (*right*) for the best five configurations obtained with our method. The array is circular, with radius $R = 75a$, where $a$ is the radius of the fibril stalk. The center-to-center spacing among fibrils is $d = 3a$, the Young's modulus of the fibrils is $E_f = E$, with $E$ the modulus of the backing layer, which has Poisson's ratio $\nu = 0.5$. The solutions are obtained using various initial configurations.

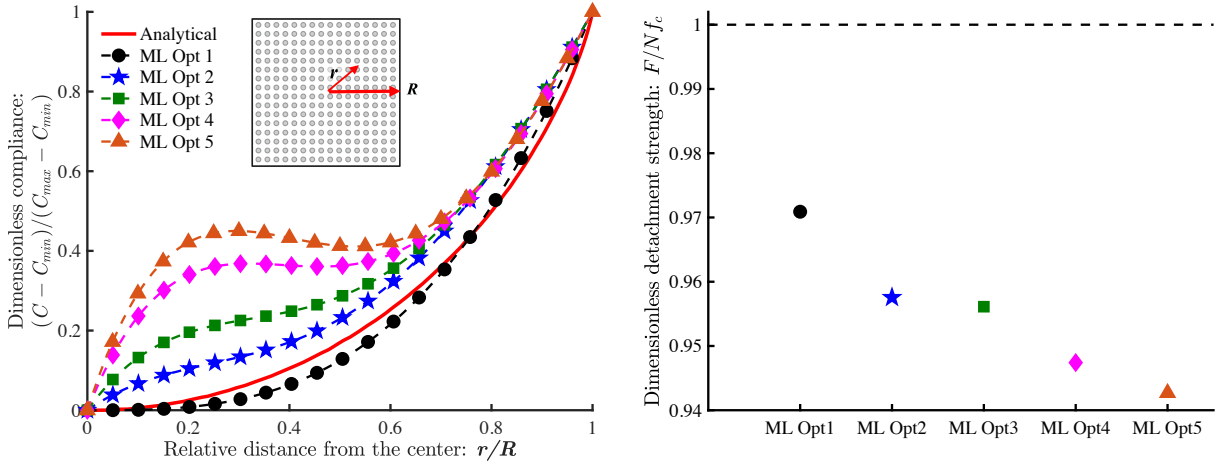

**Figure 5** Optimal fibril compliance distribution (*left*) and corresponding adhesion (detachment) strength (*right*) for the best five configurations obtained with our method. The array is square, with same outer dimension of the circle in Figure 4 (square side is equal to circle diameter). The center-to-center spacing among fibrils, as well as the elastic properties of the backing layer and the fibrils, is the same as in Figure 4. The method adopted is the same as in Figure 4.

Finally, we showcase our design tool to optimize the compliance distribution of a triangular fibrillar array and report our results in Figure 6. The circumradius of the triangle is the same as that of the square analyzed in Figure 5. We also adopt the same fibril spacing and elastic properties of BL and fibrils. In this case we don't have a theoretical benchmark for comparison on the compliance distribution from [29]. In Figure 6 we report the three best solution from our method.

We can still observe that compliant peripheral fibrils, compared to the centrally located ones, give higher adhesive strength. Unlike in the previous two cases, here we have that the biggest discrepancy among solutions lies in the compliance distribution of the peripheral fibrils rather than that of the central ones. Also, our method can achieve a maximum dimensionless adhesive strength of almost 0.93, thus a few % smaller than what was obtained in the previous cases.

Our ML-based method has proven effective in replicating the previously obtained optimal design of fibrillar adhesives, in the terms of compliance distribution generating ELS for circular and square arrays. We also obtained new results with our method, by optimizing the design of a triangular array. The simple concept of reducing the compliance of peripheral regions of an adhesive, commonly carrying the largest load, was theoretically introduced by [27,29], proven experimentally by [19], and is here confirmed via our inverse design approach. The engineer has now a new design tool to maximize adhesive strength of an interface that does not require manufacturing fibrils at the nanoscale. However, functional grading of fibril compliance represents several challenges as discussed by [29]. Fibril compliance can be tailored by adjusting the crosslink density of the elastomer composing the fibril, as well as adjusting the cross-sectional radius of the fibril. An alternative technique is adjusting the fibril length, but this will produce interfacial curvatures and challenge one's ability to create the exquisite conformation of the adhesive to the surface and create residual pre-tension in the fibrils, as explored by [31]. The exploration of the optimal design in tandem with manufacturing challenges can be a natural extension of this work, and can finally produce a comprehensive design tool for fibrillar adhesives, and can be extended to the design of functional interfaces and materials.

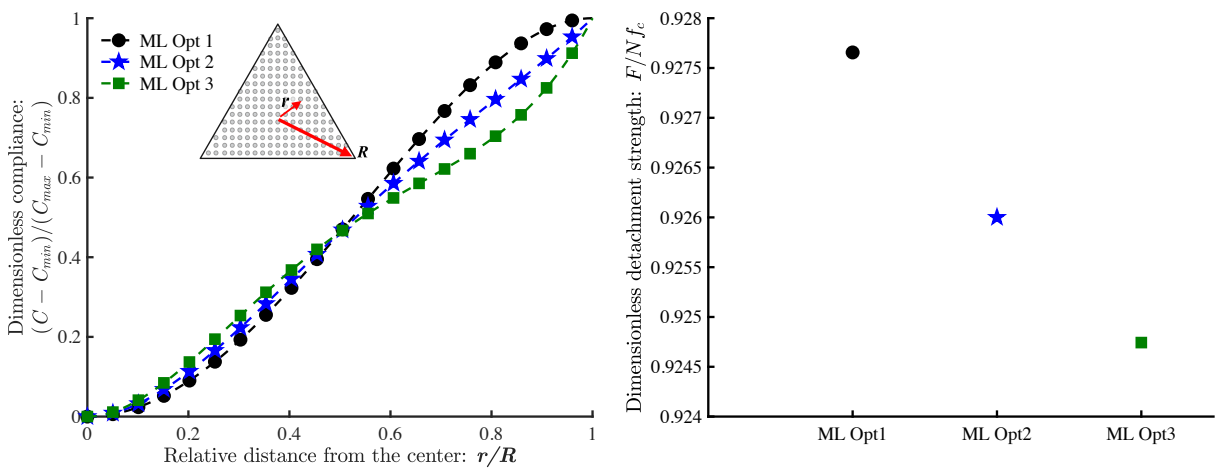

**Figure 6** Optimal compliance distribution (left figure) and the corresponding adhesion force (right figure) for fibrils in an equilateral triangle array with height of 1.5R using various initializations.

**Conclusion**

In this study we propose a gradient-based optimization method coupled with machine learning (ML) to design fibrillar adhesives by optimally distributing fibril compliance. Such optimum provides the condition of equal load sharing (ELS) in which all fibrils detach simultaneously, thereby generating the maximum theoretical adhesive strength given by the sum of the pull-off force of all fibrils. The optimum consists of more compliant fibrils at the periphery of the adhesive, as proposed by [27,29], and stiffer fibrils at the center. Such a functional grading of compliance enhances adhesion by reducing stress concentrations and was proven experimentally on lap shear joins [19]. Our optimum is compared against previous theoretical findings [29], which provided a closed-form solution of compliance distribution for circular and square adhesives. Our results align well with such a theoretical benchmark and highlight the importance of each detail in ensureing optimal adhesion. *E.g.*, we find that peripheral fibrils are more important than central ones in ensuring robust adhesion for the circular and square array. We also use our method to analyze perviously unexplored geometries such as a triangular array. In this new venture we observe the same general design principle of softer peripheral fibrils, albeit some differences emerge in the detailed distribution, and we also observe that central fibrils become more important compared to peripheral ones in this case. Our tool can then be used to further extend this study to more complex geometries in fibrullar adhesives, and can be extended to be used in other contexts, such as the design of functional interfaces and materials to provide the desired macroscopic physical behavior from the design of microstructural features.

**Conflict of Interest**
The author(s) declared no potential conflicts of interest with respect to the research, authorship, and/or publication of this article.

**Funding**
The work of MS and MB was supported by the Natural Sciences and Engineering Research Council of Canada (NSERC) (RGPIN-2017–04464). The work of AL is supported by the European Research Council (ERC) through the Starting Grant "ALPS" (101039481). Views and opinions expressed are however those of the author(s) only and do not necessarily reflect those of the European Union or the ERC Executive Agency. Neither the European Union nor the granting authority can be held responsible for them.